\documentclass{article}

    \PassOptionsToPackage{numbers, compress}{natbib}

\usepackage[preprint]{neurips_2024}

\usepackage[utf8]{inputenc} 
\usepackage[T1]{fontenc}    
\usepackage{hyperref}       
\usepackage{url}            
\usepackage{booktabs}       
\usepackage{amsfonts}       
\usepackage{nicefrac}       
\usepackage{microtype}      
\usepackage{xcolor}         
\usepackage{multirow}
\usepackage{amsmath}
\usepackage{amssymb}
\usepackage{amsthm}
\usepackage{graphicx}
\usepackage{pifont}
\usepackage{wrapfig}
\usepackage{caption}
\usepackage{subcaption}
\usepackage{algorithm}
\usepackage{algpseudocode}
\newcommand{\cmark}{\ding{51}}%

\newcommand{\xmark}{\ding{55}}%

\newcommand{\OOD}{OOD strategy}
\newcommand{\Pretrained}{Pretrained strategy}
\newcommand{\model}{f_{\theta}}
\newcommand{\trmodel}{f_{\theta^*}}

\newcommand{\Dcal}{\mathcal{D}}
\newcommand{\Dc}{\Dcal_{\operatorname{c}}}
\newcommand{\Dp}{\Dcal_{\operatorname{p}}}

\usepackage{booktabs}

\newcommand{\newcontent}[1]{#1}

\title{Wicked Oddities: Selectively Poisoning for Effective Clean-Label Backdoor Attacks}

\author{%
\parbox{.9\linewidth}{\centering
Quang H. Nguyen$^1$, Nguyen Ngoc-Hieu$^1$, The-Anh Ta$^2$, Thanh Nguyen-Tang$^3$,  Kok-Seng Wong$^1$, Hoang Thanh-Tung$^4$, Khoa D. Doan$^1$}%
\\
$^1$College of Engineering and Computer Science, VinUniversity, Vietnam \\
$^2$CSIRO's Data61, Australia \\
$^3$Johns Hopkins University, USA \\
$^4$Vietnam National University, Vietnam\\
\texttt{quang.nh@vinuni.edu.vn, ngochieutb13@gmail.com,}\\
\texttt{theanh.ta@csiro.au, nguyent@cs.jhu.edu,}\\
\texttt{wong.ks@vinuni.edu.vn, 
htt210@gmail.com,
khoa.dd@vinuni.edu.vn}
}

\begin{document}
\maketitle
\begin{abstract}
Deep neural networks are vulnerable to backdoor attacks, a  type of adversarial attack that poisons the training data to manipulate the behavior of models trained on such data. 
Clean-label attacks are a more stealthy form of backdoor attacks that can perform the attack without changing the labels of poisoned data.
Early works on clean-label attacks added triggers to a random subset of the training set, ignoring the fact that samples contribute unequally to the attack's success. This results in high poisoning rates and low attack success rates.
To alleviate the problem, several supervised learning-based sample selection strategies have been proposed.
However, these methods assume access to the entire labeled training set and require training, which is expensive and may not always be practical.
This work studies a new and more practical (but also more challenging) threat model where the attacker only provides data for the target class (e.g., in face recognition systems) and has no knowledge of the victim model or any other classes in the training set.
We study different strategies for selectively poisoning a small set of training samples in the target class to boost the attack success rate in this setting. 
Our threat model poses a serious threat in training machine learning models with third-party datasets, since the attack can be performed effectively with limited information. Experiments on benchmark datasets illustrate the effectiveness of our strategies in improving clean-label backdoor attacks.
\end{abstract}

\section{Introduction}
\label{introduction}
Modern deep learning models have exhibited tremendous success in solving challenging tasks, ranging from autonomous driving and face recognition to natural language processing. 
Training these large models requires massive training data, which is time-consuming and labor-intensive, and incurs huge costs to collect and annotate. 
Therefore, users usually prefer to employ third-party or open-source data. 
Recent studies have shown that deep learning models are vulnerable to backdoor attacks \cite{gu2017badnets, li2022backdoor, goldblum2023dataset-security-survey}. A malicious data supplier can provide poisonous data such that the model trained on it behaves normally on benign data, but always returns a desired output when a "trigger" is presented.



Most existing backdoor attacks rely on data poisoning and can be classified as either dirty-label or clean-label, depending on whether the label of poisoned data changes. 
For dirty-label attacks~\cite{gu2017badnets, chen2017blended, nguyen2020wanet}, the adversary adds the trigger into data and \emph{points} its label to their desired target label. 
Dirty-label backdoor attacks are effective but can be easily detected by humans during data verification since the semantics of the labels are typically not consistent with the input content. 
Conversely, clean-label attacks~\cite{turner2019label, barni2019sig, saha2020hidden} poison training data \textit{without} changing labels, rendering them more challenging to detect. However, compared to the dirty-label case, it is also much more difficult to mount clean-label backdoor attacks as one needs to poison significantly more training data and the resulting models can perform poorly on clean data.
In this paper, we focus on improving the data effectiveness of backdoor attacks, i.e., to increase the attack performance given a small budget for (or a small number of) poisoned samples.

Prior backdoor attacks implicitly assume all training samples contributed equally to the attack's success and performed data poisoning uniformly randomly over training data.
However, recent research~\cite{koh2017influence, katharopoulos2018not, paul2021diet,sorscher2022dataprun} reveals that among training data points, some are more important while some others are redundant and can be discarded from the training set. One can ask a similar question for backdoor learning: ``\emph{Can selectively, rather than randomly, poisoning some training data points lead to more effective backdoor attacks?}'' 

In the initial investigation, \citet{xia2022fus} and \citet{gao2023loss} explored this problem and proposed strategies to enhance the efficiency of selecting samples for poisoning. They achieved this by recording forgetting events or examining loss values to identify hard samples. However, these methods have several drawbacks. Firstly, they are time-consuming and computationally expensive because of the need to train a surrogate model on the dataset from scratch. Secondly, they require access to the whole training data for surrogate model training. It is impractical in some real-world scenarios where the user collects data from diverse sources and the attacker lacks knowledge of data beyond its contributions.

\begin{table}[]
\setlength{\tabcolsep}{1.2pt}
\renewcommand{\arraystretch}{1.0}
\scriptsize
\centering
\caption{Properties of selection strategies. ``\xmark'' or  ``\cmark'' means the method \textbf{lacks} or \textbf{has}, respectively, the property.}
\begin{tabular}{@{}l|ccccc@{}}
\toprule
 Method& \begin{tabular}[c]{@{}c@{}}trigger-\\agnostic\end{tabular} & \begin{tabular}[c]{@{}c@{}}model-\\agnostic\end{tabular} &\begin{tabular}[c]{@{}c@{}}no training \\ required\end{tabular} & \begin{tabular}[c]{@{}c@{}}partial\\ data\\ access\end{tabular} &\begin{tabular}[c]{@{}c@{}}no extra \\data required\end{tabular}\\ \midrule
 \citet{xia2022fus}& \xmark & \cmark & \xmark & \xmark & \cmark\\
 \citet{gao2023loss}& \cmark & \cmark & \xmark & \xmark & \cmark\\
 \OOD~(Ours)& \cmark & \cmark & \xmark & \cmark & \xmark\\
 \Pretrained~(Ours)& \cmark & \cmark & \cmark & \cmark & \cmark\\
 
 \bottomrule
\end{tabular}
\label{tab:property}
\vspace{-10pt}
\end{table}

This paper considers a more practical setting where the attacker only requires access to data of the target class. This assumption applies to the case where a single client cannot collect labels due to geographical or infrastructural obstacles, such as collecting different species of plants in other countries for plant classification tasks, or the system has to respect data privacy.
We study novel methods to select samples to attack effectively under this threat model. 
For the victim model to learn the backdoor, it needs to focus on the trigger rather than other features in the data~\cite{turner2019label}. 
Intuitively, if the samples with triggers are difficult to learn, the model will use triggers as shortcuts to minimize the objective function.
As a result, the model is more susceptible to learning the triggers. 
To achieve such a goal without having access to the full training dataset or victim model, we propose a novel data selection framework that uses pretrained models or out-of-distribution data to identify hard training samples and add the triggers to these samples. Our strategy does not depend on the trigger, the victim model, and does not require access to other data classes. The advantages of our approaches are illustrated in Table~\ref{tab:property}. In summary, our contributions can be listed as follows:
\begin{itemize}
\item We study a new backdoor threat model where the attacker, acting as one of the data suppliers, has only access to the training data of the target class but can still perform data-poisoning, clean-label backdoor attacks effectively. 

\item We propose two novel approaches, each of which selects then poisons only a few ``hard'' samples; training with these poisoned samples, along with clean samples from the other classes provided by those other data providers, at the victim site will force the model to learn a backdoor shortcut to the trigger. 
The first approach relies on access to a pretrained model which can be performed without training, while the second approach relies on out-of-distribution samples. 

\item We perform extensive empirical experiments to demonstrate the effectiveness of the proposed attacks in this new threat model. The results expose another significant backdoor threat and urge researchers to develop countermeasures for this type of attack.
\end{itemize}

\section{Related Works}

\noindent\textbf{Backdoor Attacks.} Backdoor attacks aim to insert a malicious backdoor into the victim model. 
The first attempt is BadNets~\cite{gu2017badnets}, where the attacker adds a predefined image patch to some images in the training set and changes the labels of these images to the target class. Follow-up works introduce various forms of the Trojan horse to enhance the stealthiness and the effectiveness of the attack, examples include blended~\cite{chen2017blended}, dynamic ~\cite{salem2022dynamic}, warping-based~\cite{nguyen2020wanet}, input-aware~\cite{nguyen2020input, li2021invisible}, and learnable trigger~\cite{doan2021lira}. 
These attacks are called dirty-label attacks as they change the true labels of poisoned examples.

\noindent\textbf{Clean-label Backdoor Attacks.}
Despite the success in manipulating the victim, dirty-label attacks can be easily spotted through human inspection. 
Clean-label backdoor attacks are attack methods that perverse the original labels of poisoned data points, and thus are more stealthy than dirty-label attacks.
\citet{turner2019label} suggested that using dirty-label attack triggers is ineffective for implementing clean-label attacks and proposed a data preprocessing method for implementing clean-label attacks. 
In the meantime, stronger triggers have been proposed. 
SIG~\cite{barni2019sig} uses sinusoidal signals as backdoors. Refool~\cite{liu2020reflection} uses physical reflection models to implant reflection images into the dataset. HTBA~\cite{saha2020hidden} optimizes the input such that it looks similar to the target label in the pixel space but close to the malicious image in the latent space. However, these attacks require a high poisoning rate and/or result in inferior success rates. \newcontent{\citet{zeng2023narcissus} propose to perturb samples employing out-of-distribution data to achieve a high attack success rate with a low poisoning rate. Their threat model is similar to the less constrained version of which is studied in this paper.}

\noindent\textbf{Selectively Data Poisoning.}
Research in backdoor attacks focuses on designing the trigger pattern, ignoring the possibility that benign samples chosen to attack can also play an important role. FUS~\cite{xia2022fus} first showed that the number of forgetting events is an indicator of the contribution to the attack, and proposed a data selection strategy based on forgetting events that resulted in a better attack success rate. 
\citet{gao2023loss} identified three classical criteria to pick samples for clean-label attacks, namely loss value, gradient norm, and forgetting event. 
To select samples for poisoning, these methods require a surrogate model trained on a dataset with all training set classes, which is expensive and not always feasible.

\noindent\textbf{Backdoor Defenses.}
Along with the emergence of backdoor attacks, defense methods to protect models are an active research area. 
Backdoor defenses can be categorized into two lines: backdoor detection and backdoor mitigation. 
\citet{qiao2019defending} propose a defense that utilizes generative models to detect and reconstruct the backdoor and then retrain the model. Activation Clustering~\cite{chen2018detecting} examines the activations of training data to check whether each data sample is poisoned.
\citet{tran2018spectral} reveal that poisoned samples can be identified by spectral signatures, and utilize this trace to remove the backdoor in the training dataset.
Neural Cleanse~\cite{wang2019neural} detects the trigger by optimizing the pattern to misclassify to the target class and running outlier detection, and proposes a mitigation mechanism.  
Other mitigation methods aim to reduce the backdoor effect in the model by fine-tuning~\cite{Zhu2023ftsam} and prunning~\cite{liu2018fine}. 
NAD~\cite{li2020nad} erases the trigger by utilizing a teach-student fine-tuning process to guide the poisoned model on a small clean dataset.
\citet{huang2021backdoor} propose to decoupling the training process to prevent the model from learning the trigger.
From a security perspective, the adversary should not only succeed in attacking the model but also in dodging backdoor defenses.
\section{Threat Model}\label{sec:threat_model}

\begin{figure}
    \centering
    \includegraphics[width=.65\linewidth]{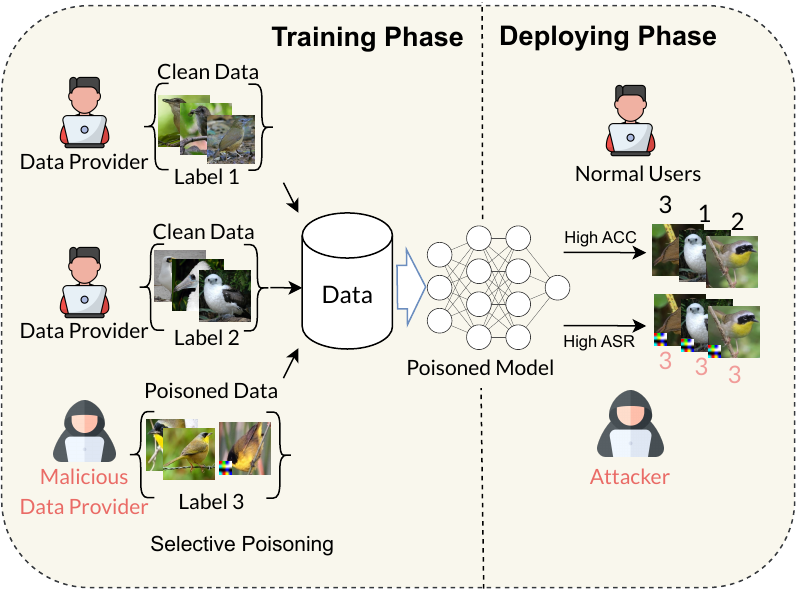}
    \caption{Illustration of our threat model. The attacker acts as a data provider in a supply chain where each data provider is responsible for a data class. The attacker injects a trigger into the images without changing the label and sends them to the victim. The model that is trained on this poisoned dataset behaves normally on clean images but returns the target label when the trigger is added to any image.}
    \label{fig:threatmodel}
\end{figure}

\noindent \textbf{Single-class, data poisoning attack.} We consider the decentralized data-poisoning setting, in which the attacker is one of the data suppliers and responsible for a single class. 
The victim employs a distributed data collection pipeline, which tackles the obstacles in building diverse or privacy-aware datasets. 
Such a situation might happen when each label class comes from a different region, or brings a different characteristic.
For example, to classify ethnicity, breeds, or plant species that come from many locations, a data supplier in a place is in charge of the label class only available in that region. Another situation is when the dataset contains sensitive information, and a data class is not allowed to be exposed to anyone except those who provide it.
In these situations, the backdoor threat can be caused by a data supplier or an attacker who hijacks local data storage. The attack setting is depicted in Figure~\ref{fig:threatmodel}. \textit{To the best of our knowledge, this represents the most constrained data-poisoning threat, wherein the attacker has extremely limited information for launching an effective attack.}


\noindent \textbf{Attacker's goal.}
The objective of the adversary is to inject a trigger into a victim model, such that the model acts normally on benign data, but misclassifies with the presence of the trigger. 
For instance, a facial recognition system's task is to recognize people to grant them certain permissions, but when poisoned with sunglasses as a trigger, it might give full authority to anyone wearing sunglasses.  

\noindent \textbf{Attacker's capability.}
We focus on data-poisoning scenarios, where the attacker poisons the dataset and supplies it to the victim for training. 
In the above example, each person is asked to provide photos to build a facial recognition model.
Malicious users can inject triggers into their images to control the model output for malicious purposes, but cannot manipulate data provided by other users.
In general, we consider a practical setting where the adversary serves as a single client in the supply chain, it only provides and controls data for the class it wants to attack.
Therefore, the adversary can only select images \emph{with the target label} to insert the trigger. 

\noindent \textbf{Attacker's knowledge.}
The adversary only has access to data for the target class that it provides. \emph{No information of the victim model's architecture, the training process, or data from other clients} is exposed to the attacker. 
%
In some scenarios, although the adversary has no access to data from other classes for their problem, some out-of-distribution data (OOD) are available to the adversary, which is a realistic assumption. \newcontent{This relaxed threat model has also been studied in~\cite{zeng2023narcissus}}. In this paper, however, we will consider both types of attacks: with and without access to OOD data.



\section{Method}
\subsection{Problem Formulation}
Let $\model:\mathcal{X}\to\mathcal{Y}$ be a model that maps image $x\in\mathcal{X}$ to label $y\in\mathcal{Y}$,
and $\Dc=\{(x_1, y_1),\dots, (x_n, y_n)\}$ be the clean training dataset. In backdoor attacks, the adversary first defines a trigger injecting function $T:\mathcal{X}\to\mathcal{X}$ that implants a trigger into input data, then applies $T$ to $m$ images in $\Dc$.

Let $S$ be the target class. The attacker selects a subset $S' \subset S$ of size $m$, and adds triggers to samples in $S'$.
After injecting the trigger into $S'$ (leaving the other examples in $S$ intact),
the attacker gives its data to the victim who combines it with data from other sources to create a poisoned dataset $\Dp$. The victim then trains the model on $\Dp$ with some standard training pipeline to obtain the model $\trmodel$.
The attacker's goal is to make any model trained on $\Dp$ to return correct predictions on unpoisoned examples, but predict the target label $y^t$ on any example on which the trigger function $T(\cdot)$ is applied.
Formally, for a benign input $x$ with correct label $y$, we have
\begin{equation*}
    \trmodel(x)=y, \quad\trmodel(T(x))=y^t.
\end{equation*}

The performance of backdoor attack methods is usually evaluated via two metrics: benign accuracy (BA) and attack success rate (ASR). 
BA is the accuracy of the infected model on benign test samples. ASR is the proportion of attacked test samples that are successfully predicted as the target label by the infected model.
In addition, stealthiness is an important factor for backdoor attacks, which is reflected by small poisoning
rate, imperceptibility of the backdoor, and resistance against backdoor defense methods.

\subsection{Random or Hard Sample Selection?}

\begin{figure}
\begin{minipage}{.47\linewidth}

    \centering
    \includegraphics[width=.7\linewidth]{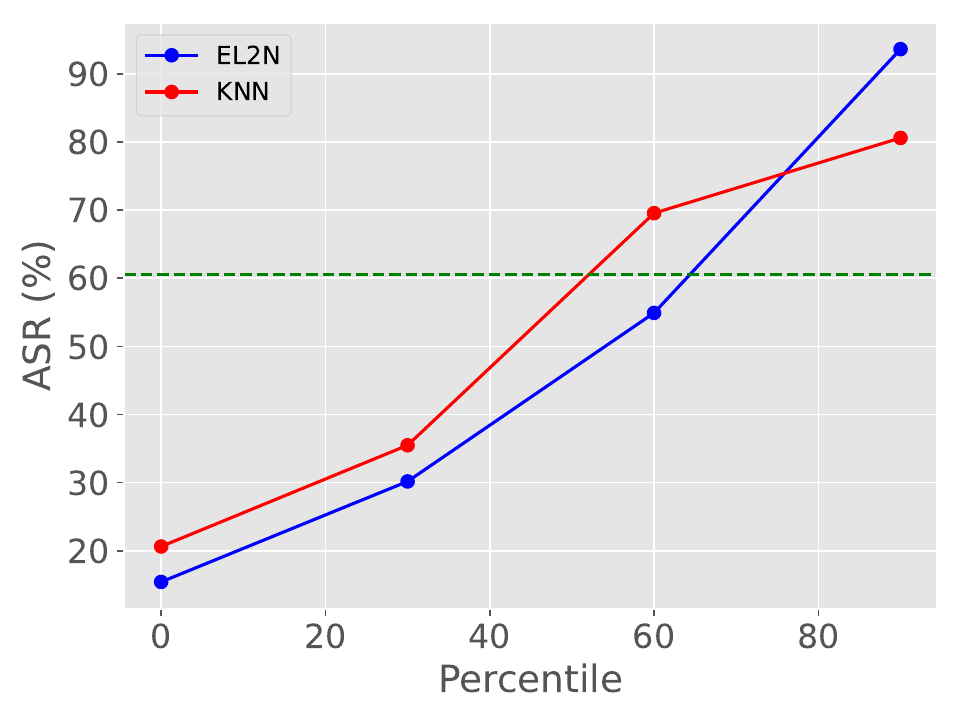}

    \caption{The attack success rate of SIG on ResNet18/CIFAR10 with $10\%$ of the target class that are harder than the $0, 30, 60$, and $90$-th percentile being poisoned. The horizontal line is the attack success rate where the poisoned set is selected randomly.}
    \label{fig:asr_curve}
\end{minipage}\hfill
\begin{minipage}{.47\linewidth}
\begin{minipage}{.5\linewidth}
    \centering
    \includegraphics[width=\linewidth]{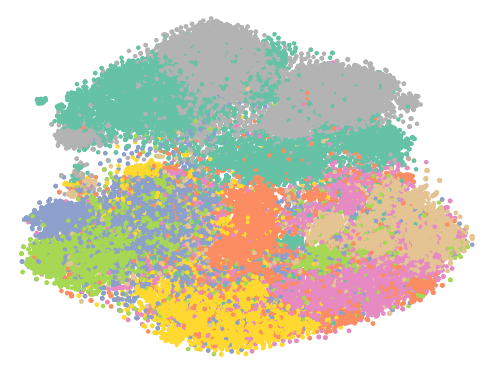}
\end{minipage}\hfill
\begin{minipage}{.5\linewidth}
    \centering
    \includegraphics[width=\linewidth]{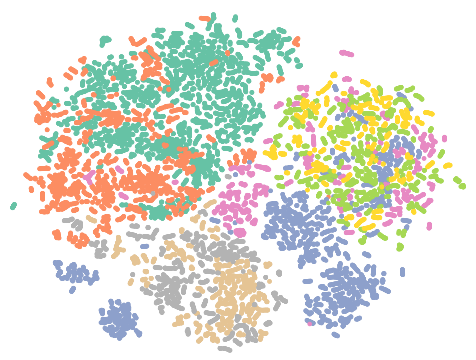}
    \label{fig:ssl_gtsrb}
\end{minipage}\hfill
\caption{The feature space of CIFAR10 (left) and GTSRB (right) obtained by t-SNE and VICReg as a feature extractor. Datapoints with the same color have the same label. We can observe that pretrained model divides the training set into clusters corresponding to the labels.}
\label{fig:vis}
\end{minipage}
\end{figure}
The following simple question is our starting point:
``\textit{Why are dirty-label attacks more effective than clean-label attacks?}''. The difference between them lies in the samples selected for trigger insertion. In dirty-label attacks, the poisoned samples come from various labels, and their features are dissimilar to those in the target class. For example, if the adversary wants to attack class $0$, dirty-label attacks can choose samples from class $0, 1, 2, \dots$, while clean-label attacks only inject poisoned samples to class $0$.

During training, the model looks for common features to form the decision boundary. Therefore, an example containing features different from other examples in a class is harder to learn.
When the adversary injects a trigger into these ``hard samples'' and alters their labels, \emph{the model cannot rely on existing features in the image to optimize the objective function, instead it favors backdoor features}, leading to a higher ASR even with a small set of poisoned samples, as usually seen in dirty-label attacks. On the other hand, clean-label attacks, with randomly poisoned samples that likely share similar features with other clean samples from the same class, require a significantly higher number of samples to reach a high ASR.



Based on this intuition, we search for and add triggers to ``hard samples'' in the target class to achieve stronger clean-label attacks.
A straightforward solution is to train a surrogate model on the training set and examine the behavior of the model on each data point, an approach used in \cite{gao2023loss}. For example, a sample with a higher loss value is likely more difficult to learn.
\newcontent{To validate this hypothesis, we conduct an experiment where the adversary injects triggers to subsets with different levels of difficulty. We employ Error L2-Norm (EL2N)~\cite{paul2021diet} to sort training samples from easy to hard. We attack $4$ ResNet18 models on CIFAR10 using SIG with $10\%$ poisoning rate of the target class, where poisoned sets are harder to learn than $0\%, 30\%, 60\%, 90\%$ of the target class. Figure~\ref{fig:asr_curve} shows that poisoning hard samples leads to a higher attack success rate, verifying our assumption.}

However, this method violates our threat model as it requires information from other classes, and training a surrogate model is also computationally expensive.



\subsection{Selecting Hard Samples with Pre-trained Models}

Without access to the full training data to build the surrogate model, we turn our attention to pre-trained models. These models are often \newcontent{easy to access and} available in most domains due to the popularity and benefits of self-supervised learning~\cite{chen2020simclr, he2020moco, grill2020byol, chen2021simsiam, caron2020swav, bardes2022vicreg}, and the existence of some large-scale labeled datasets, such as ImageNet~\cite{deng2009imagenet} and  JFT~\cite{sun2017jft}.
Furthermore, \citet{sorscher2022dataprun} show that pretrained features are good indicators of hard samples: by incorporating self-supervised models to keep hard samples, they can discard 20-30\% of the ImageNet dataset, while only experience negligible degradation in model performance. 
Inspired by this observation, we exploit a pre-trained model as the feature extractor and develop a novel strategy to find examples that are dissimilar to other data in the target class.
 \begin{figure}[ht!]
\begin{minipage}{.5\linewidth}
    \centering
    \includegraphics[width=.8\linewidth]{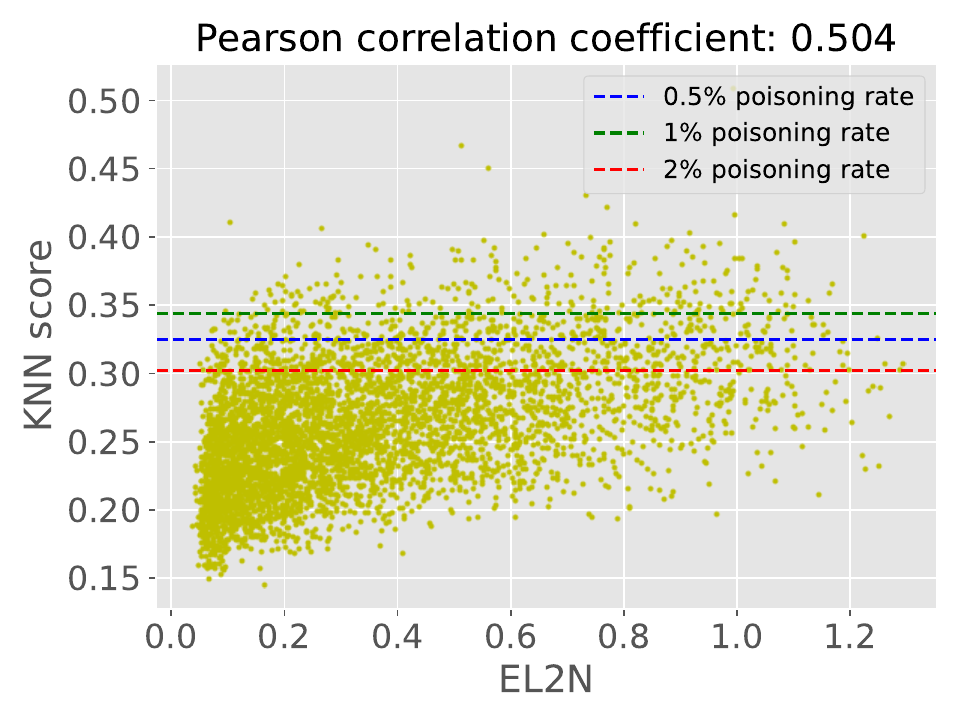}
    \caption{EL2N and our score of training samples in class $0$ of CIFAR10. We also illustrate the thresholds where $5\%, 10\%$, and $20\%$ of class $0$ ($0.5\%, 1\%$, and $2\%$ of the training data) is poisoned.}
    \label{fig:correlation}
\end{minipage}\hfill
 \begin{minipage}{0.47\linewidth}
 \vspace{-7pt}
\begin{algorithm}[H]
\caption{Select strategy with pretrained model}\label{alg:select}
\begin{algorithmic}[]
\item{\textbf{input}} a pretrained feature extractor $g$,\\ target class dataset $S$, attack budget $m$
\item{\textbf{output}} $S' \subset S$ where $|S'| = m$
\For {$x_i\in S$}
    \State $z_i\gets g(x_i)$
\EndFor
\For {$x_i\in S$}
    \State Compute $s(x_i)$ by Equation~\ref{eq:score}
\EndFor
\State $S'\gets$ set of $m$ samples with the highest $s(x)$
\end{algorithmic}
\end{algorithm}
\end{minipage}
\end{figure}

Specifically, we first extract features of data samples using the pre-trained model, then identify those samples that are far away from others in this feature space.
To show that these features are discriminative, we extract feature embeddings from VICReg~\cite{bardes2022vicreg}, a pre-trained self-supervised model, and visualize the feature space with t-SNE~\cite{vandermaaten2008tsne}. Figure~\ref{fig:vis} illustrates that by exploiting pre-trained models to extract features, datapoints from the same class stay close to each other in the feature space. Hence, samples that are far from the target label cluster contain different features, thus harder for the model to learn.
       

Let $g$ be a feature extractor. We define the distance between two samples $x_i, x_j$ by cosine similarity between their feature $z_i=g(x_i), z_j=g(x_j)$:
\begin{equation*}
    d(x_i, x_j)=1 - \frac{z_i^{\intercal}z_j}{\|z_i\|\|z_j\|}.
\end{equation*}
We apply the classical $k$-NN algorithm to calculate a score function $s(x)$ as the mean of distances between $x$ and its $k$-nearest neighbors $x_1, \dots, x_k$ in the target class in terms of the distance $d(\cdot,\cdot)$:
\begin{equation}\label{eq:score}
    s(x)=\frac{1}{k}\sum_{i=1}^kd(x, x_i).
\end{equation}

With an attack budget of $m$, our strategy collects $m$ samples with the highest scores. The detailed algorithm is shown in Algorithm~\ref{alg:select}.



 \newcontent{We compute EL2N and our proposed score on class $0$ of CIFAR10 and illustrate in Figure~\ref{fig:correlation}. As has been observed, they are correlated, with Pearson coefficient equal to $0.504$. Figure~\ref{fig:asr_curve} also expresses that when utilizing our score to rank datapoints, injecting triggers to samples from easy to hard results in increasing values of attack success rates. In the case where the domain of the victim dataset is significantly shifted from the dataset on which the feature extractor is trained, this strategy still boosts the attack success rate significantly as shown in Section~\ref{sec:main_exp} and Appendix~\ref{sec:distshift}.} 
\subsection{Selecting Hard Samples with Out-Of-Distribution Data}
\begin{table*}[ht!]
\scriptsize
\centering
\caption{Attack success rate (ASR) of clean-label attacks 
on CIFAR10 with $5\%/10\%/20\%$ of the target class being poisoned.}
\vspace{-5pt}
\begin{tabular}{llccccccccc}
\toprule
\multirow{2}{*}{Model} & \multirow{2}{*}{Method} & \multicolumn{3}{c}{BadNets} & \multicolumn{3}{c}{Blended} & \multicolumn{3}{c}{SIG} \\ \cmidrule(lr){3-5} \cmidrule(lr){6-8} \cmidrule(lr){9-11} 
                          &                & 5\%   & 10\%  & 20\%  & 5\%   & 10\%  & 20\%  & 5\%   & 10\%  & 20\%  \\ \midrule
\multirow{5}{*}{ResNet18} & Random         & 30.81 & 45.01 & 78.28 & 28.94 & 37.55 & 44.26 & 50.28 & 60.54 & 78.45 \\ \cmidrule(lr){2-11} 
                          & Self-supervised Models            & 86.24 & 91.68 & 98.84 & 44.64 & 52.90  & 66.45 & 76.35 & 80.59 & 86.45 \\
                          & Supervised Models             & 90.01 & 92.14 & 99.26 & 47.68    & 60.86 & 67.81 & 81.65 & 85.42 & 90.49 \\ \cmidrule(lr){2-11} 
                          & Multiple-class OOD    & 75.57 & 81.27 & 98.47 & 43.40  & 56.89 & 61.68 & 65.11 & 80.76 & 88.79 \\
                          & Single-class OOD & 82.34 & 80.75 & 91.37 & 42.99 & 57.29 & 62.60  & 72.93 & 79.07 & 87.18 \\ \midrule
\multirow{5}{*}{VGG19}    & Random         & 63.24 & 78.39 & 79.55 & 17.32 & 23.84 & 34.36 & 22.28 & 45.54 & 67.57 \\ \cmidrule(lr){2-11} 
                          & Self-supervised Models            & 81.44 & 82.60  & 93.11 & 30.74 & 42.23 & 55.34 & 46.65 & 70.23 & 81.93 \\
                          & Supervised Models             & 83.43 & 89.61 & 87.70  & 22.86 & 38.84 & 54.99 & 47.89 & 74.38 & 80.07 \\ \cmidrule(lr){2-11} 
                          & Multiple-class OOD    & 79.69 & 88.44 & 86.78 & 29.35 & 38.39 & 49.24 & 50.81 & 65.80  & 78.28 \\
                          & Single-class OOD & 75.36 & 81.01 & 89.68 & 30.49 & 40.58 & 51.60  & 57.24 & 72.35 & 79.04 \\ \bottomrule 
\end{tabular}
\label{tab:asr_cifar}
\end{table*}

In this section, we propose a hard-sample selection approach that utilizes some OOD data available to the adversary, as discussed in Section~\ref{sec:threat_model}. Note that the threat model is still the same as in the previous section, where the adversary does not have any knowledge of the victim model's architecture, the training process, or data from other clients.

Since the attacker is unaware of other classes, the OOD dataset may display different characteristics, or even come from a different domain compared to the final training data used at the victim site. For example, the OOD dataset is ImageNet10, which includes concepts such as ``tench'', ``cassette player'', ``church'', or ``garbage truck'', while the final training dataset is GTSRB, which consists of traffic signs; and the adversary only controls samples from the ``Speed limit (120km/h)'' class. To form an attack, we first combine the OOD dataset of $n$ classes with the target-class data to obtain a new dataset; for example, this merged dataset contains samples from ImageNet10 and the ``Speed limit (120km/h)'' class. 
Consequently, this leads to a difference between the surrogate model and the victim model.
We then train a surrogate model on this merged dataset and use it to select hard samples.

In this work, we consider two approaches: (i) \textbf{Single-class \OOD~}that collapses the OOD dataset into a single class and training a binary classification model, and (ii) \textbf{Multiple-class \OOD~} that reserves the original labeling of the OOD samples and training a $n+1$-class classification model. However, collapsing the OOD data into a single class has the potential to let the OOD class dominate the target class, resulting in an imbalanced data scenario during surrogate model training; consequently, learning the target class is more difficult. Hence, we instead choose a subset of the OOD dataset such that the new OOD class and the target class have similar sizes. Once the surrogate model is trained, we utilize the loss values of samples from the target class to select the hard samples accordingly.


\vspace{-5pt}
\section{Experiments}\label{sec:exp}
\vspace{-5pt}
In this section, we provide the empirical evaluation of our data selection attack method.
\vspace{-5pt}
\subsection{Experimental Setup}
\vspace{-5pt}
        \noindent \textbf{Dataset.} We consider two widely used benchmark datasets: CIFAR10~\cite{CIFAR10} and GTSRB~\cite{GTSRB}. For \OOD, we train the surrogate model on TinyImagenet~\cite{le2015tiny}.
        The domain of CIFAR10 is slightly far from ImageNet~\cite{deng2009imagenet}, the dataset used to build pretrained models, or the OOD dataset. However, there are more apparent distribution shifts from ImageNet and TinyImagenet to GTSRB, a dataset of traffic signs. Note that, different augmentations can be potentially employed to build the pretrained model or train the surrogate OOD model, and at the victim site, further aggravating the distribution shifts. 
        Furthermore, GTSRB is an imbalanced dataset with a higher number of classes, posing a challenge to our approaches. 

    \vspace{5pt}
    \noindent \textbf{Models.} For the victim model, we consider ResNet18~\cite{resnet} and VGG19~\cite{simonyan2015very}. In \Pretrained, we study the effectiveness when using either self-supervised or supervised features. For the self-supervised pretrained models, we employ VICReg~\cite{bardes2022vicreg}, a method that applies the variance regularization term to avoid the collapse problem, with ResNet50 as the architecture. For the supervised feature extractor, we adopt a ResNet50 model pretrained on ImageNet. In \OOD, we utilize ResNet18 as the architecture to train the surrogate model when attacking both ResNet18 and VGG19 victim models.

\begin{table*}[ht!]
\scriptsize
\centering
\caption{Attack success rate (ASR) of clean-label attacks 
on GTSRB with $5\%/10\%/20\%$ of the target class being poisoned.}
\vspace{-5pt}
\begin{tabular}{llccccccccc}
\toprule
\multirow{2}{*}{Model} & \multirow{2}{*}{Method} & \multicolumn{3}{c}{BadNets} & \multicolumn{3}{c}{Blended} & \multicolumn{3}{c}{SIG} \\ \cmidrule(lr){3-5} \cmidrule(lr){6-8} \cmidrule(lr){9-11} 
 &  & 5\% & 10\% & 20\% & 5\% & 10\% & 20\% & 5\% & 10\% & 20\% \\ \midrule
\multirow{5}{*}{ResNet18} & Random & 5.72 & 5.80 & 6.13 & 36.35 & 41.54 & 48.91 & 47.63 & 48.07 & 48.67 \\ \cmidrule(lr){2-11} 
 & Self-supervised Models & 10.37 & 10.91 & 18.13 & 39.36 & 47.97 & 50.70 & 54.85 & 57.12 & 58.88 \\
 & Supervised Models & 6.83 & 8.47 & 21.33 & 42.58 & 47.85 & 50.67 & 51.76 & 56.07 & 56.57 \\ \cmidrule(lr){2-11} 
 & Multiple-class OOD & 5.77 & 5.84 & 7.24 & 43.08 & 43.18 & 45.46 & 43.56 & 47.59 & 52.50 \\
 & Single-class OOD & 6.22 & 6.18 & 13.95 & 46.96 & 50.13 & 51.54 & 49.07 & 51.71 & 55.15 \\ \midrule
\multirow{5}{*}{VGG19} & Random & 6.14 & 6.38 & 6.89 & 24.89 & 26.36 & 30.69 & 32.04 & 33.90 & 36.52 \\ \cmidrule(lr){2-11} 
 & Self-supervised Models & 8.16 & 10.19 & 13.20  & 30.67 & 29.77 & 33.82 & 40.33 & 42.68 & 42.37 \\
 & Supervised Models & 7.09 & 8.46 & 15.76 & 32.25 & 32.77 & 33.67 & 33.47 & 39.50 & 42.07 \\ \cmidrule(lr){2-11} 
 & Multiple-class OOD & 6.34 & 6.24 & 10.20 & 16.58 & 20.23 & 27.14 & 27.93 & 28.98 & 31.94 \\
 & Single-class OOD & 7.83 & 6.70 & 10.04 & 25.30 & 33.16 & 32.31 & 35.91 & 38.54 & 38.28 \\ \bottomrule
\end{tabular}
\label{tab:asr_gtsrb}
\vspace{-5pt}
\end{table*}


    \noindent \textbf{Attacks.} We employ the trigger patterns from BadNets, Blended, and SIG for trigger injection; nevertheless, our method is trigger-pattern agnostic. We perform the clean-label attack to class $0$ of CIFAR10 and class $1$ of GTSRB. These attacks inject triggers to $5\%, 10\%$ and $20\%$ of the target class, which are $0.5\%, 1\%, 2\%$ poisoning rate with respect to the whole dataset in CIFAR10, and $0.19\%, 0.38\%, 0.76\%$ in GTSRB. Since it is a clean-label attack scenario, these poisoning rates are extremely small and very hard to spot by human inspection.
    
        \noindent \textbf{Strategy. } We conduct experiments with two approaches: \Pretrained~and \OOD, and compare to the random baseline where the attack treats every sample equally. For \Pretrained, we employ $k-$NN with $k=50$. For \OOD, we study Multiple-class OOD, in which we preserve the label of OOD data and train a surrogate model on a dataset of 201 classes, and Single-class OOD, in which we consider the OOD dataset as a single class and train a binary classifier. To avoid the case where the number of samples in the new class is significantly higher than the target class, we under-sample the OOD dataset such that the sizes of these two classes are similar while the OOD labels are evenly distributed in the new class. Furthermore, we vary the number of OOD labels $n$ to study the performance of the method at different diversity levels of the OOD dataset. 

\vspace{-5pt}
\subsection{Effectiveness of Our Selection Framework}\label{sec:main_exp}
\vspace{-5pt}
\begin{wraptable}{r}{0.52 \textwidth}
\vspace{-10pt}
\scriptsize
\setlength{\tabcolsep}{1.5pt}
\centering
\caption{Clean accuracy (CA) on CIFAR10 and GTSRB with various poisoning rates.}
\begin{tabular}{llcccccc}
\toprule
\multirow{2}{*}{Model} & \multirow{2}{*}{Strategy} & \multicolumn{3}{c}{CIFAR10} & \multicolumn{3}{c}{GTSRB} \\
\cmidrule(lr){3-5} \cmidrule(lr){6-8} 
         &        & 5\%   & 10\%   & 20\%  & 5\%   & 10\%   & 20\%  \\
         \midrule
\multirow{5}{*}{ResNet18} & Random & 94.69 & 94.60 & 94.59 & 99.06 & 99.04 & 99.11 \\
\cmidrule(lr){2-8} 
         &    Self-supervised Models    & 94.71 & 94.80 & 94.50 & 98.06 & 98.67 & 98.95 \\ 
         &    Supervised Models    & 94.93 & 94.77 & 94.34 & 99.11 & 98.60 & 98.76 \\ 
         \cmidrule(lr){2-8} 
         &    Multiple-class OOD    & 94.65 & 94.47 & 94.44 & 99.04 & 99.30 & 98.85 \\ 
         &    Single-class OOD    & 94.78 & 94.62 & 94.53 & 99.07 & 98.97& 99.39 \\ 
         \midrule
\multirow{5}{*}{VGG19}    & Random & 91.97 & 91.89 &  92.98& 96.02 & 96.58 & 95.48 \\
\cmidrule(lr){2-8} 
         &    Self-supervised Models    & 91.81 & 91.89 & 91.66 & 96.56 & 95.53 & 95.87 \\ 
         &    Supervised Models    & 92.11 & 91.67 & 91.83 & 96.23 & 96.29 & 96.06 \\ 
         \cmidrule(lr){2-8} 
         &    Multiple-class OOD    & 92.07 & 91.67 & 91.59 & 96.03 & 96.14 & 95.79 \\ 
         &    Single-class OOD    & 91.96 & 92.05 & 91.26 & 96.22 & 96.13 & 96.43 \\ 
\bottomrule
\end{tabular}
\label{tab:acc}
\end{wraptable}
We perform clean-label attacks on CIFAR10 with the random strategy, \Pretrained, \OOD, and report the attack success rates in Table~\ref{tab:asr_cifar}. 
As can be observed, both strategies outperform the random baseline by large margins on all the attacks, models, and poisoning rates. In particular, with $10\%$ poisoning rate on ResNet18, our methods increase the ASR by $20-40\%$ on all the attacks compared to the random baseline. Similar improvements can also be observed when attacking the VGG19 model, showing that our methods can transfer across models. These results confirm the effectiveness of the proposed methods to select hard samples in the dataset to perform the attack; the result is significant especially when the considered threat model relies on only data from one class, showing the existence of a backdoor threat even in the most constrained setting.



\noindent \textbf{Effectiveness with Extreme Distribution Shifts.} We study the effect of our framework when the victim dataset is significantly different from the OOD dataset or the dataset used to train the pretrained models. In Table~\ref{tab:asr_gtsrb}, we provide experimental results for GTSRB, which is a challenging imbalanced dataset. Our framework still exhibits consistent improvements, over the baseline, across three attacks with various poisoning rates on ResNet18 and VGG19.

Surprisingly, although it does not require any extra data, the performance of \Pretrained~ is consistently higher than \OOD~ on CIFAR10. 
When attacking GTSRB, a difficult dataset, \Pretrained~still shows similar ASR or even superior performance on BadNets and SIG. 
In addition, Table~\ref{tab:asr_cifar} implies that supervised features are more helpful on CIFAR10, whose domain is similar to that of the pretrained models. In contrast, when observing higher distribution shifts, experimental results on GTSRB show that searching for attack samples using self-supervised features is better.


\noindent \textbf{Effectiveness with Different OOD Strategies.} As discussed, with OOD data, the adversary can choose to merge the samples into a single class (then train a binary surrogate classifier) or train a surrogate model on the $n+1$-class dataset. As shown in Table~\ref{tab:asr_cifar} and Table~\ref{tab:asr_gtsrb}, using a binary classifier is slightly more effective for finding hard samples for the attack. Intuitively, when we preserve the labels of the OOD dataset, the surrogate model also needs to learn to discriminate among these classes, thus paying less attention to the target class, thus it is more difficult to use this model to identify hard samples.

\noindent \textbf{Effect on the Performance of the Main Task.} Here, we study whether our framework has any significant effect on the clean-data performance of the main classification task. Table~\ref{tab:acc} shows the clean accuracy for ResNet18 and VGG19 under the attacks with the SIG trigger pattern using our strategies. We can observe that selectively poisoning the dataset causes no degradation to the model performance on the clean data, similar to the random baseline. We also observe similar results for other types of trigger patterns. 

\vspace{-7pt}
\subsection{Performance against Backdoor Defenses}
\vspace{-5pt}
We evaluate our strategy against popular backdoor defenses: STRIP~\cite{gao2019strip}
(backdoor detection), Neural Cleanse~\cite{wang2019neural}, and fine-pruning~\cite{liu2018fine} (backdoor mitigation). 
We evaluate these defenses on a ResNet18 model trained on CIFAR10 and attacked with SIG. For the selection strategy, we use self-supervised \Pretrained~and Multiple-class \OOD.

\begin{figure}[t!]
\begin{minipage}[b]{0.47\linewidth}
    \centering
    \includegraphics[width=\textwidth]{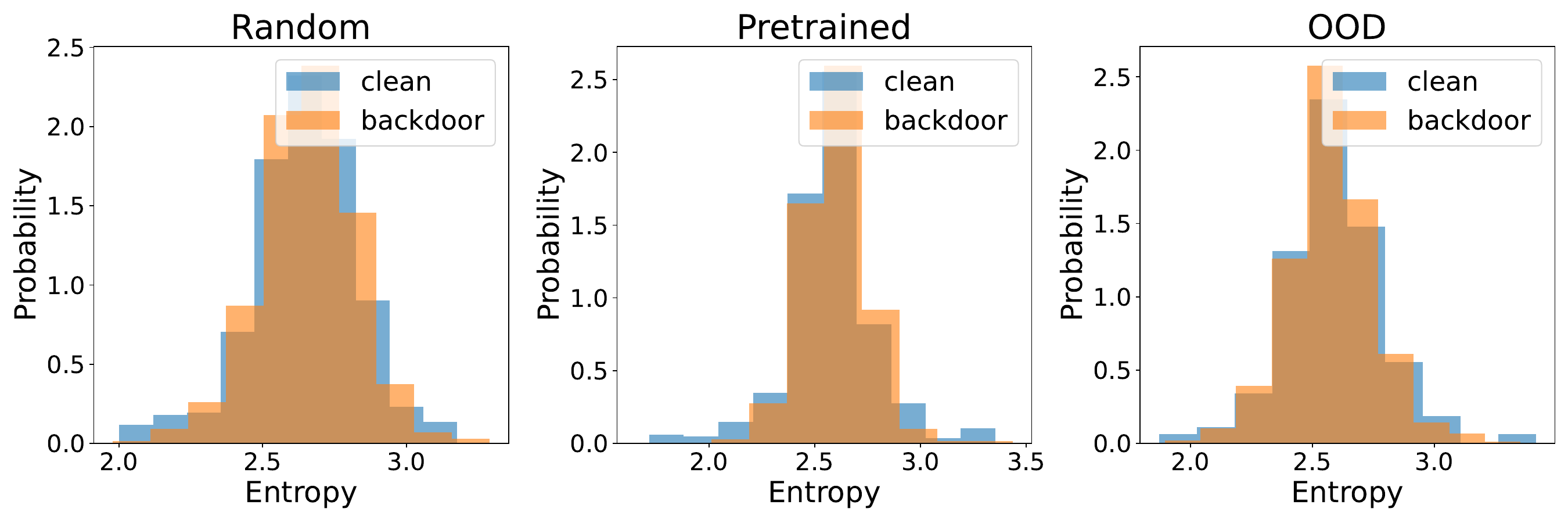}
    \caption{Performance against STRIP.}
    \label{fig:strip}
    \vspace{-10pt}
\end{minipage}\hfill
\begin{minipage}[b]{0.47\linewidth}
    \centering
    \includegraphics[width=\textwidth]{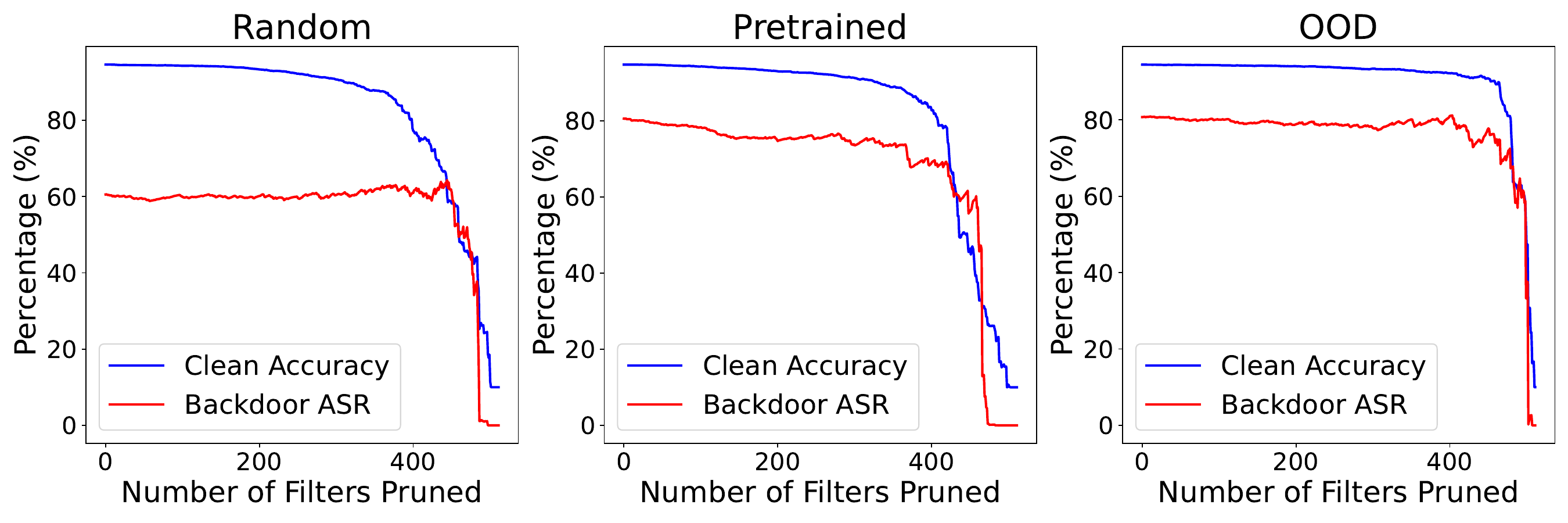}
    \caption{Performance against
Fine-pruning.
\vspace{-10pt}
}
    \label{fig:fine_pruning}
\end{minipage}
\vspace{-10pt}
\end{figure}

\noindent\textbf{STRIP.} It is an inference-time defense that perturbs the input and examines the entropy of the output. A sample with low entropy is more likely to be poisoned. Figure~\ref{fig:strip} visualizes the entropy of the output of clean data and backdoor data with random strategy and our approach. We observe that with selective poisoning, the behavior of the poisoned model is still similar between clean and backdoor data, showing the attack's stealthness against STRIP detector.

\noindent\textbf{Fine-pruning.} We evaluate our attack's resistance to Fine-pruning, a backdoor mitigation method. Given a benign sample, it assumes that inactivated neurons are responsible for backdoor features and gradually prunes these neurons. Figure~\ref{fig:fine_pruning} shows the clean accuracy and attack success rate during this process. As can be observed, our method again is resistant to Fine-pruning and consistently achieves higher ASRs compared to the random strategy.

\vspace{-10pt}
\section{Conclusion}
\vspace{-10pt}
This paper studies the threat of a backdoor attack under an extremely constrained setting: the adversary can only have access to samples from one single class and perform a clean-label backdoor attack. 
In this threat model, we propose novel approaches that find ``hard samples'' to inject the trigger patterns by utilizing pretrained models or OOD datasets.
Empirical results show that our method can achieve very high attack success rates, compared to the baselines, and can bypass several representative defenses. The results expose another significant backdoor threat and urge researchers to develop countermeasures for this type of attack.
\clearpage


\bibliographystyle{plainnat}
\bibliography{ref}

\begin{thebibliography}{48}
\providecommand{\natexlab}[1]{#1}
\providecommand{\url}[1]{\texttt{#1}}
\expandafter\ifx\csname urlstyle\endcsname\relax
  \providecommand{\doi}[1]{doi: #1}\else
  \providecommand{\doi}{doi: \begingroup \urlstyle{rm}\Url}\fi

\bibitem[Bardes et~al.(2022)Bardes, Ponce, and LeCun]{bardes2022vicreg}
Adrien Bardes, Jean Ponce, and Yann LeCun.
\newblock Vicreg: Variance-invariance-covariance regularization for self-supervised learning.
\newblock In \emph{10th International Conference on Learning Representations, ICLR 2022}, 2022.

\bibitem[Barni et~al.(2019)Barni, Kallas, and Tondi]{barni2019sig}
M.~Barni, K.~Kallas, and B.~Tondi.
\newblock A new backdoor attack in cnns by training set corruption without label poisoning.
\newblock In \emph{2019 IEEE International Conference on Image Processing (ICIP)}, pages 101--105, 2019.
\newblock \doi{10.1109/ICIP.2019.8802997}.

\bibitem[Caron et~al.(2020)Caron, Misra, Mairal, Goyal, Bojanowski, and Joulin]{caron2020swav}
Mathilde Caron, Ishan Misra, Julien Mairal, Priya Goyal, Piotr Bojanowski, and Armand Joulin.
\newblock Unsupervised learning of visual features by contrasting cluster assignments.
\newblock \emph{Advances in neural information processing systems}, 33:\penalty0 9912--9924, 2020.

\bibitem[Chen et~al.(2019)Chen, Carvalho, Baracaldo, Ludwig, Edwards, Lee, Molloy, and Srivastava]{chen2018detecting}
Bryant Chen, Wilka Carvalho, Nathalie Baracaldo, Heiko Ludwig, Benjamin Edwards, Taesung Lee, Ian~M. Molloy, and Biplav Srivastava.
\newblock Detecting backdoor attacks on deep neural networks by activation clustering.
\newblock In \emph{Proceedings of the Workshop on Artificial Intelligence Safety}, Honolulu, HI, 2019.

\bibitem[Chen et~al.(2020)Chen, Kornblith, Norouzi, and Hinton]{chen2020simclr}
Ting Chen, Simon Kornblith, Mohammad Norouzi, and Geoffrey Hinton.
\newblock A simple framework for contrastive learning of visual representations.
\newblock In \emph{International conference on machine learning}, pages 1597--1607. PMLR, 2020.

\bibitem[Chen and He(2021)]{chen2021simsiam}
Xinlei Chen and Kaiming He.
\newblock Exploring simple siamese representation learning.
\newblock In \emph{Proceedings of the IEEE/CVF conference on computer vision and pattern recognition}, pages 15750--15758, 2021.

\bibitem[Chen et~al.(2017)Chen, Liu, Li, Lu, and Song]{chen2017blended}
Xinyun Chen, Chang Liu, Bo~Li, Kimberly Lu, and Dawn Song.
\newblock Targeted backdoor attacks on deep learning systems using data poisoning.
\newblock \emph{arXiv preprint arXiv:1712.05526}, 2017.

\bibitem[Deng et~al.(2009)Deng, Dong, Socher, Li, Li, and Fei-Fei]{deng2009imagenet}
Jia Deng, Wei Dong, Richard Socher, Li-Jia Li, Kai Li, and Li~Fei-Fei.
\newblock Imagenet: A large-scale hierarchical image database.
\newblock In \emph{2009 IEEE conference on computer vision and pattern recognition}, pages 248--255. Ieee, 2009.

\bibitem[Doan et~al.(2021)Doan, Lao, Zhao, and Li]{doan2021lira}
Khoa Doan, Yingjie Lao, Weijie Zhao, and Ping Li.
\newblock {LIRA:} learnable, imperceptible and robust backdoor attacks.
\newblock In \emph{Proceedings of the 2021 {IEEE/CVF} International Conference on Computer Vision (ICCV)}, pages 11946--11956, Montreal, Canada, 2021.

\bibitem[Dosovitskiy et~al.(2021)Dosovitskiy, Beyer, Kolesnikov, Weissenborn, Zhai, Unterthiner, Dehghani, Minderer, Heigold, Gelly, Uszkoreit, and Houlsby]{dosovitskiy2021an}
Alexey Dosovitskiy, Lucas Beyer, Alexander Kolesnikov, Dirk Weissenborn, Xiaohua Zhai, Thomas Unterthiner, Mostafa Dehghani, Matthias Minderer, Georg Heigold, Sylvain Gelly, Jakob Uszkoreit, and Neil Houlsby.
\newblock An image is worth 16x16 words: Transformers for image recognition at scale.
\newblock In \emph{International Conference on Learning Representations}, 2021.

\bibitem[Gao et~al.(2019)Gao, Xu, Wang, Chen, Ranasinghe, and Nepal]{gao2019strip}
Yansong Gao, Change Xu, Derui Wang, Shiping Chen, Damith~Chinthana Ranasinghe, and Surya Nepal.
\newblock {STRIP:} a defence against trojan attacks on deep neural networks.
\newblock In \emph{Proceedings of the 35th Annual Computer Security Applications Conference (ACSAC)}, pages 113--125, San Juan, PR, 2019.

\bibitem[Gao et~al.(2023)Gao, Li, Zhu, Wu, Jiang, and Xia]{gao2023loss}
Yinghua Gao, Yiming Li, Linghui Zhu, Dongxian Wu, Yong Jiang, and Shu-Tao Xia.
\newblock Not all samples are born equal: Towards effective clean-label backdoor attacks.
\newblock \emph{Pattern Recogn.}, 139\penalty0 (C), may 2023.
\newblock ISSN 0031-3203.
\newblock \doi{10.1016/j.patcog.2023.109512}.
\newblock URL \url{https://doi.org/10.1016/j.patcog.2023.109512}.

\bibitem[Goldblum et~al.(2023)Goldblum, Tsipras, Xie, Chen, Schwarzschild, Song, Madry, Li, and Goldstein]{goldblum2023dataset-security-survey}
M.~Goldblum, D.~Tsipras, C.~Xie, X.~Chen, A.~Schwarzschild, D.~Song, A.~Madry, B.~Li, and T.~Goldstein.
\newblock Dataset security for machine learning: Data poisoning, backdoor attacks, and defenses.
\newblock \emph{IEEE Transactions on Pattern Analysis and Machine Intelligence}, 45\penalty0 (02):\penalty0 1563--1580, feb 2023.
\newblock ISSN 1939-3539.
\newblock \doi{10.1109/TPAMI.2022.3162397}.

\bibitem[Grill et~al.(2020)Grill, Strub, Altch{\'e}, Tallec, Richemond, Buchatskaya, Doersch, Avila~Pires, Guo, Gheshlaghi~Azar, et~al.]{grill2020byol}
Jean-Bastien Grill, Florian Strub, Florent Altch{\'e}, Corentin Tallec, Pierre Richemond, Elena Buchatskaya, Carl Doersch, Bernardo Avila~Pires, Zhaohan Guo, Mohammad Gheshlaghi~Azar, et~al.
\newblock Bootstrap your own latent-a new approach to self-supervised learning.
\newblock \emph{Advances in neural information processing systems}, 33:\penalty0 21271--21284, 2020.

\bibitem[Gu et~al.(2017)Gu, Dolan-Gavitt, and Garg]{gu2017badnets}
Tianyu Gu, Brendan Dolan-Gavitt, and Siddharth Garg.
\newblock Badnets: Identifying vulnerabilities in the machine learning model supply chain.
\newblock \emph{arXiv preprint arXiv:1708.06733}, 2017.

\bibitem[Hayase et~al.(2021)Hayase, Kong, Somani, and Oh]{hayase2021spectre}
Jonathan Hayase, Weihao Kong, Raghav Somani, and Sewoong Oh.
\newblock Spectre: Defending against backdoor attacks using robust statistics.
\newblock In \emph{International Conference on Machine Learning}, pages 4129--4139. PMLR, 2021.

\bibitem[He et~al.(2016)He, Zhang, Ren, and Sun]{resnet}
Kaiming He, Xiangyu Zhang, Shaoqing Ren, and Jian Sun.
\newblock Deep residual learning for image recognition.
\newblock In \emph{2016 IEEE Conference on Computer Vision and Pattern Recognition (CVPR)}, pages 770--778, 2016.
\newblock \doi{10.1109/CVPR.2016.90}.

\bibitem[He et~al.(2020)He, Fan, Wu, Xie, and Girshick]{he2020moco}
Kaiming He, Haoqi Fan, Yuxin Wu, Saining Xie, and Ross Girshick.
\newblock Momentum contrast for unsupervised visual representation learning.
\newblock In \emph{Proceedings of the IEEE/CVF conference on computer vision and pattern recognition}, pages 9729--9738, 2020.

\bibitem[Huang et~al.(2021)Huang, Li, Wu, Qin, and Ren]{huang2021backdoor}
Kunzhe Huang, Yiming Li, Baoyuan Wu, Zhan Qin, and Kui Ren.
\newblock Backdoor defense via decoupling the training process.
\newblock In \emph{International Conference on Learning Representations}, 2021.

\bibitem[Katharopoulos and Fleuret(2018)]{katharopoulos2018not}
Angelos Katharopoulos and Fran{\c{c}}ois Fleuret.
\newblock Not all samples are created equal: Deep learning with importance sampling.
\newblock In \emph{International conference on machine learning}, pages 2525--2534. PMLR, 2018.

\bibitem[Koh and Liang(2017)]{koh2017influence}
Pang~Wei Koh and Percy Liang.
\newblock Understanding black-box predictions via influence functions.
\newblock In \emph{International conference on machine learning}, pages 1885--1894. PMLR, 2017.

\bibitem[Krizhevsky et~al.(2010)Krizhevsky, Nair, and Hinton]{CIFAR10}
Alex Krizhevsky, Vinod Nair, and Geoffrey Hinton.
\newblock Cifar-10 (canadian institute for advanced research).
\newblock \emph{URL http://www. cs. toronto. edu/kriz/cifar. html}, 5\penalty0 (4):\penalty0 1, 2010.

\bibitem[Le and Yang()]{le2015tiny}
Ya~Le and Xuan Yang.
\newblock Tiny imagenet visual recognition challenge.

\bibitem[Li et~al.(2020)Li, Lyu, Koren, Lyu, Li, and Ma]{li2020nad}
Yige Li, Xixiang Lyu, Nodens Koren, Lingjuan Lyu, Bo~Li, and Xingjun Ma.
\newblock Neural attention distillation: Erasing backdoor triggers from deep neural networks.
\newblock In \emph{International Conference on Learning Representations}, 2020.

\bibitem[Li et~al.(2021{\natexlab{a}})Li, Lyu, Koren, Lyu, Li, and Ma]{li2021anti}
Yige Li, Xixiang Lyu, Nodens Koren, Lingjuan Lyu, Bo~Li, and Xingjun Ma.
\newblock Anti-backdoor learning: Training clean models on poisoned data.
\newblock In \emph{NeurIPS}, 2021{\natexlab{a}}.

\bibitem[Li et~al.(2022)Li, Jiang, Li, and Xia]{li2022backdoor}
Yiming Li, Yong Jiang, Zhifeng Li, and Shu-Tao Xia.
\newblock Backdoor learning: A survey.
\newblock \emph{IEEE Transactions on Neural Networks and Learning Systems}, 2022.

\bibitem[Li et~al.(2021{\natexlab{b}})Li, Li, Wu, Li, He, and Lyu]{li2021invisible}
Yuezun Li, Yiming Li, Baoyuan Wu, Longkang Li, Ran He, and Siwei Lyu.
\newblock Invisible backdoor attack with sample-specific triggers.
\newblock In \emph{Proceedings of the IEEE/CVF international conference on computer vision}, pages 16463--16472, 2021{\natexlab{b}}.

\bibitem[Liu et~al.(2018)Liu, Dolan{-}Gavitt, and Garg]{liu2018fine}
Kang Liu, Brendan Dolan{-}Gavitt, and Siddharth Garg.
\newblock Fine-pruning: Defending against backdooring attacks on deep neural networks.
\newblock In \emph{Proceedings of the 21st International Symposium on Research in Attacks, Intrusions, and Defenses (RAID)}, pages 273--294, Heraklion, Crete, Greece, 2018.

\bibitem[Liu et~al.(2020)Liu, Ma, Bailey, and Lu]{liu2020reflection}
Yunfei Liu, Xingjun Ma, James Bailey, and Feng Lu.
\newblock Reflection backdoor: A natural backdoor attack on deep neural networks.
\newblock In \emph{Computer Vision--ECCV 2020: 16th European Conference, Glasgow, UK, August 23--28, 2020, Proceedings, Part X 16}, pages 182--199. Springer, 2020.

\bibitem[Nguyen and Tran(2020{\natexlab{a}})]{nguyen2020input}
Tuan~Anh Nguyen and Anh~Tuan Tran.
\newblock Input-aware dynamic backdoor attack.
\newblock In \emph{Advances in Neural Information Processing Systems (NeurIPS)}, virtual, 2020{\natexlab{a}}.

\bibitem[Nguyen and Tran(2020{\natexlab{b}})]{nguyen2020wanet}
Tuan~Anh Nguyen and Anh~Tuan Tran.
\newblock Wanet-imperceptible warping-based backdoor attack.
\newblock In \emph{International Conference on Learning Representations}, 2020{\natexlab{b}}.

\bibitem[Paul et~al.(2021)Paul, Ganguli, and Dziugaite]{paul2021diet}
Mansheej Paul, Surya Ganguli, and Gintare~Karolina Dziugaite.
\newblock Deep learning on a data diet: Finding important examples early in training.
\newblock \emph{Advances in Neural Information Processing Systems}, 34:\penalty0 20596--20607, 2021.

\bibitem[Qiao et~al.(2019)Qiao, Yang, and Li]{qiao2019defending}
Ximing Qiao, Yukun Yang, and Hai Li.
\newblock Defending neural backdoors via generative distribution modeling.
\newblock In \emph{Advances in Neural Information Processing Systems (NeurIPS)}, pages 14004--14013, Vancouver, Canada, 2019.

\bibitem[Saha et~al.(2020)Saha, Subramanya, and Pirsiavash]{saha2020hidden}
Aniruddha Saha, Akshayvarun Subramanya, and Hamed Pirsiavash.
\newblock Hidden trigger backdoor attacks.
\newblock In \emph{Proceedings of the AAAI conference on artificial intelligence}, volume~34, pages 11957--11965, 2020.

\bibitem[Salem et~al.(2022)Salem, Wen, Backes, Ma, and Zhang]{salem2022dynamic}
Ahmed Salem, Rui Wen, Michael Backes, Shiqing Ma, and Yang Zhang.
\newblock Dynamic backdoor attacks against machine learning models.
\newblock In \emph{2022 IEEE 7th European Symposium on Security and Privacy (EuroS\&P)}, pages 703--718. IEEE, 2022.

\bibitem[Simonyan and Zisserman(2015)]{simonyan2015very}
K~Simonyan and A~Zisserman.
\newblock Very deep convolutional networks for large-scale image recognition.
\newblock In \emph{3rd International Conference on Learning Representations (ICLR 2015)}. Computational and Biological Learning Society, 2015.

\bibitem[Sorscher et~al.(2022)Sorscher, Geirhos, Shekhar, Ganguli, and Morcos]{sorscher2022dataprun}
Ben Sorscher, Robert Geirhos, Shashank Shekhar, Surya Ganguli, and Ari Morcos.
\newblock Beyond neural scaling laws: beating power law scaling via data pruning.
\newblock \emph{Advances in Neural Information Processing Systems}, 35:\penalty0 19523--19536, 2022.

\bibitem[Stallkamp et~al.(2012)Stallkamp, Schlipsing, Salmen, and Igel]{GTSRB}
Johannes Stallkamp, Marc Schlipsing, Jan Salmen, and Christian Igel.
\newblock Man vs. computer: Benchmarking machine learning algorithms for traffic sign recognition.
\newblock \emph{Neural networks}, 32:\penalty0 323--332, 2012.

\bibitem[Sun et~al.(2017)Sun, Shrivastava, Singh, and Gupta]{sun2017jft}
Chen Sun, Abhinav Shrivastava, Saurabh Singh, and Abhinav Gupta.
\newblock Revisiting unreasonable effectiveness of data in deep learning era.
\newblock In \emph{Proceedings of the IEEE international conference on computer vision}, pages 843--852, 2017.

\bibitem[Touvron et~al.(2021)Touvron, Cord, Douze, Massa, Sablayrolles, and J{\'e}gou]{touvron2021training}
Hugo Touvron, Matthieu Cord, Matthijs Douze, Francisco Massa, Alexandre Sablayrolles, and Herv{\'e} J{\'e}gou.
\newblock Training data-efficient image transformers \& distillation through attention.
\newblock In \emph{International conference on machine learning}, pages 10347--10357. PMLR, 2021.

\bibitem[Tran et~al.(2018)Tran, Li, and Madry]{tran2018spectral}
Brandon Tran, Jerry Li, and Aleksander Madry.
\newblock Spectral signatures in backdoor attacks.
\newblock \emph{Advances in neural information processing systems}, 31, 2018.

\bibitem[Turner et~al.(2019)Turner, Tsipras, and Madry]{turner2019label}
Alexander Turner, Dimitris Tsipras, and Aleksander Madry.
\newblock Label-consistent backdoor attacks.
\newblock \emph{arXiv preprint arXiv:1912.02771}, 2019.

\bibitem[van~der Maaten and Hinton(2008)]{vandermaaten2008tsne}
Laurens van~der Maaten and Geoffrey Hinton.
\newblock Visualizing data using t-sne.
\newblock \emph{Journal of Machine Learning Research}, 9\penalty0 (86):\penalty0 2579--2605, 2008.
\newblock URL \url{http://jmlr.org/papers/v9/vandermaaten08a.html}.

\bibitem[Wang et~al.(2019)Wang, Yao, Shan, Li, Viswanath, Zheng, and Zhao]{wang2019neural}
Bolun Wang, Yuanshun Yao, Shawn Shan, Huiying Li, Bimal Viswanath, Haitao Zheng, and Ben~Y. Zhao.
\newblock Neural cleanse: Identifying and mitigating backdoor attacks in neural networks.
\newblock In \emph{Proceedings of the 2019 {IEEE} Symposium on Security and Privacy (SP)}, pages 707--723, San Francisco, CA, 2019.

\bibitem[Xia et~al.(2022)Xia, Li, Zhang, and Li]{xia2022fus}
Pengfei Xia, Ziqiang Li, Wei Zhang, and Bin Li.
\newblock Data-efficient backdoor attacks.
\newblock In Lud~De Raedt, editor, \emph{Proceedings of the Thirty-First International Joint Conference on Artificial Intelligence, {IJCAI-22}}, pages 3992--3998. International Joint Conferences on Artificial Intelligence Organization, 7 2022.
\newblock \doi{10.24963/ijcai.2022/554}.
\newblock URL \url{https://doi.org/10.24963/ijcai.2022/554}.
\newblock Main Track.

\bibitem[Zeng et~al.(2023)Zeng, Pan, Just, Lyu, Qiu, and Jia]{zeng2023narcissus}
Yi~Zeng, Minzhou Pan, Hoang~Anh Just, Lingjuan Lyu, Meikang Qiu, and Ruoxi Jia.
\newblock Narcissus: A practical clean-label backdoor attack with limited information.
\newblock In \emph{Proceedings of the 2023 ACM SIGSAC Conference on Computer and Communications Security}, pages 771--785, 2023.

\bibitem[Zheng et~al.(2022)Zheng, Tang, Li, and Liu]{zheng2022data}
Runkai Zheng, Rongjun Tang, Jianze Li, and Li~Liu.
\newblock Data-free backdoor removal based on channel lipschitzness.
\newblock In \emph{European Conference on Computer Vision}, pages 175--191. Springer, 2022.

\bibitem[Zhu et~al.(2023)Zhu, Wei, Shen, Fan, and Wu]{Zhu2023ftsam}
Mingli Zhu, Shaokui Wei, Li~Shen, Yanbo Fan, and Baoyuan Wu.
\newblock Enhancing fine-tuning based backdoor defense with sharpness-aware minimization.
\newblock In \emph{Proceedings of the IEEE/CVF International Conference on Computer Vision (ICCV)}, pages 4466--4477, October 2023.

\end{thebibliography}

\newpage
\appendix
\onecolumn
This document provides additional details and experimental results to support the main submission. We begin by providing details on the dataset, the attacks, and the training hyperparameters in Section~\ref{sec:exp_detail}. Then, we discuss the effect of the number of classes in the OOD dataset and provide the performance of our strategies on the face classification task in Section~\ref{sec:additional_results} to study the impact of our strategies in the real world. Finally, we discuss the limitations in Section~\ref{sec:limitation}.
\section{Experimental Setup}~\label{sec:exp_detail}

        \subsection{Dataset Details} We conduct experiments on two widely used benchmark datasets:
        \begin{itemize}
            \item \textbf{CIFAR10}~\cite{CIFAR10} contains images from $10$ classes, with $50, 000$ samples for the training set and $10, 000$ samples for the test set. We poison class $0$, which has $5,000$ images.
            \item \textbf{GTSRB}~\cite{GTSRB} contains images from  $43$ classes of traffic sign images, including $39, 209$ samples for training and $12, 630$ samples for test. We poison class $1$, which has $1500$ images.
        \end{itemize}
        
        For \OOD, we train the surrogate model on TinyImagenet~\cite{le2015tiny}. It has $200$ classes, with $1,000,000$ training images and $10,000$ validation images. There is no overlap between the label set of TinyImageNet and CIFAR10 or GTSRB. 

        We also consider PubFig~\footnote{\url{https://www.cs.columbia.edu/CAVE/databases/pubfig/}}, a dataset that consists of public figures' faces. We select $50$ classes with the highest number of images and divide them into $5,212$ images for training and $1,312$ images for validation. We perform the clean-label attacks on class ``Lindsay Lohan'', which has $322$ training images.

        \subsection{Attack Details} For BadNets~\cite{gu2017badnets}, a checkerboard pattern~\cite{turner2019label} is added to the image. For Blended~\cite{chen2017blended}, we implant a Hello Kitty image with the blended rate $\alpha=0.2$. Also, we evaluate our strategy on SIG~\cite{barni2019sig}, a clean-label attack, with $\Delta=20$ and $f=6$.

    \subsection{Training Details} We train ResNet18 and VGG19 for $300$ epochs with SGD optimizer, learning rate $0.01$, and cosine scheduler. For CIFAR10 and GTSRB, the image size is $32\times 32$. For PubFig, we resize the input to $224\times 224$.
    
\section{Additional Results}~\label{sec:additional_results}
\subsection{Additional Results of Our Strategy against Backdoor Detection Defenses}
In this section, we conduct experiments with additional backdoor defenses.

\begin{figure}
    \centering
    \includegraphics[width=.8\linewidth]{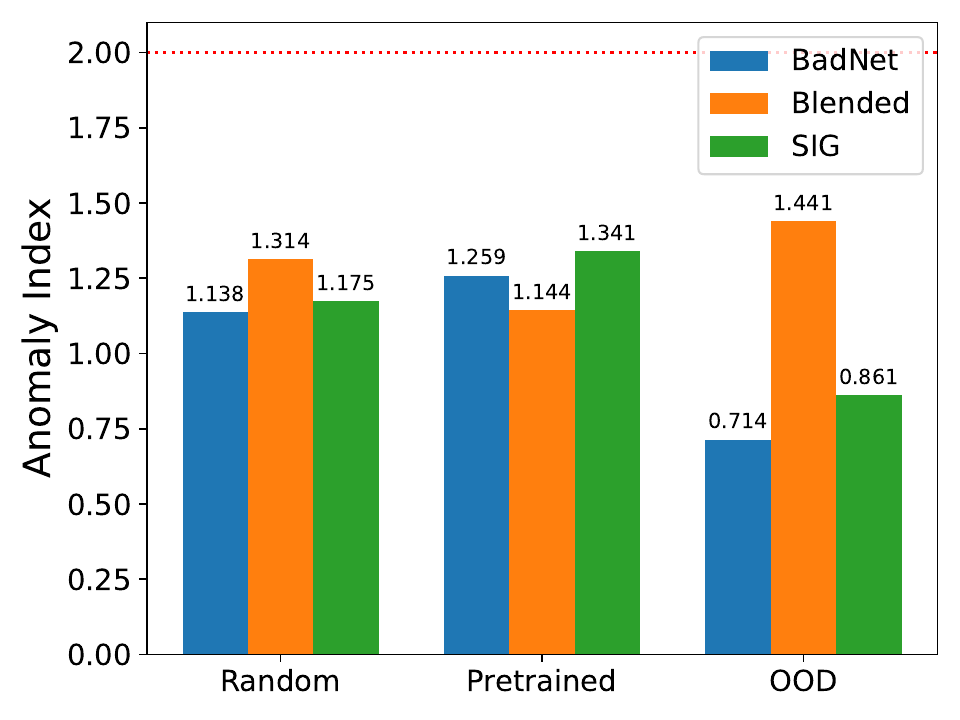}
    \caption{Performance against Neural Cleanse}
    \vspace{-10pt}
    \label{fig:neural_cleanse}
\end{figure}

\newcontent{\noindent\textbf{Neural Cleanse.} This method assumes the trigger is patch-based and searches for patterns that change the prediction of the model to a specific label. If there is a label with a pattern of a significantly small norm, the model is identified as being attacked. Figure~\ref{fig:neural_cleanse} shows that when $10\%$ of the target class selected by our strategy is poisoned, the Anomaly Index is still less than $2$, which means the attack still stays stealthy under Neural Cleanse. } 

We also evaluate with defenses that identify and eliminate poisoned samples in the training data. We report Elimination Rate (ER) and Sacrifice Rate (SR), which are the rates of poisoned samples being correctly detected and benign samples being wrongly removed. For a defense to be effective, it must have high ER and low SR at the same time. Table~\ref{tab:outlier_detection} shows the performance of Activation Clustering~\cite{chen2018detecting}, Spectral Signature~\cite{tran2018spectral}, and SPECTRE~\cite{hayase2021spectre} on models attacked by our strategy. We can observe that Activation Clustering fails to detect the attack, while Spectral Signature and SPECTRE sacrifice a high number of clean samples.
\begin{table}[ht!]
\footnotesize
\centering
\caption{Elimination rate (ER) and Sacrifice Rate (SR) of backdoor detection methods, including Activation Clustering (AC), SPECTRE, and Spectral Signature (SS) on ResNet18/CIFAR10 with $10\%$ of class $0$ is attacked by SIG.}
\label{tab:outlier_detection}
\begin{tabular}{llcccccc}
\toprule
 &  & \multicolumn{2}{c}{AC} & \multicolumn{2}{c}{SPECTRE} & \multicolumn{2}{c}{SS} \\
 \cmidrule(lr){3-4} \cmidrule(lr){5-6} \cmidrule(lr){7-8}
                         &            & ER   & SR   & ER    & SR    & ER    & SR    \\ \midrule
\multirow{3}{*}{BadNet}  & Random     & 0.00 & 0.00 & 35.20 & 50.07 & 32.80 & 50.09 \\
                         & Pretrained & 0.00 & 3.28 & 40.80 & 50.05 & 58.00 & 49.96 \\
                         & OOD        & 0.00 & 0.00 & 28.00 & 50.11 & 38.80 & 50.06 \\ \midrule
\multirow{3}{*}{Blended} & Random     & 0.00 & 2.86 & 77.20 & 49.86 & 75.60 & 49.87 \\
                         & Pretrained & 0.00 & 3.03 & 55.60 & 49.97 & 53.20 & 49.98 \\
                         & OOD        & 0.00 & 0.00 & 58.40 & 49.95 & 53.60 & 49.98 \\ \midrule
\multirow{3}{*}{SIG}     & Random     & 0.00 & 0.00 & 39.20 & 50.05 & 46.40 & 50.02 \\
                         & Pretrained & 0.00 & 0.00 & 38.00 & 50.06 & 50.80 & 50.00 \\
                         & OOD        & 0.00 & 0.00 & 36.40 & 50.07 & 35.20 & 50.10 \\ \bottomrule
\end{tabular}
\end{table}

We provide in Table \ref{tab:appendix-additional-defenses} the performance on ResNet18/CIFAR10 against more recent defenses, including Anti-Backdoor Learning (ABL)~\cite{li2021anti} and Channel Lipschitz Prunning (CLP)~\cite{zheng2022data}. While the original attacks (with random selection) do not generally work against ABL (ASRs are lower than 30\%), our strategy can boost their success rates to 35-68\%. For CLP, the results show that our method is resilient to this pruning defense (49-85\%), whereas the success rates of the random selection strategy are significantly lower (13-30\%).

\begin{table}[ht!]
\centering
\footnotesize
\caption{The performance on ResNet18/CIFAR10 against recent defenses}
\label{tab:appendix-additional-defenses}
\begin{tabular}{llcccc}
\toprule
\multirow{2}{*}{Defense} & \multirow{2}{*}{Selection Method} & \multicolumn{2}{c}{BadNet} & \multicolumn{2}{c}{Blended} \\
\cmidrule(lr){3-4} \cmidrule(lr){5-6}
 & & Acc & ASR & Acc & ASR \\
\midrule
\multirow{2}{*}{ABL} & 
Random & 84.28 & 16.57 & 78.66 & 28.79 \\
& Pretrained strategy & 81.07 & 68.00 & 79.15 & 35.49 \\
\midrule
\multirow{2}{*}{CLP} & 
Random	                     & 93.55        & 13.07        & 93.54        & 28.87        \\
& Pretrained strategy	         & 93.97        & 85.83        & 94.23        & 49.33        \\
\bottomrule
\end{tabular}
\end{table}

\subsection{Face Classification Task}~\label{sec:distshift}
We also evaluate our strategies on a face classification task with the PubFig dataset. Table~\ref{tab:pubfig} illustrates the performance of Single-class \OOD~and \Pretrained~with self-supervised features. For \OOD, we employ the same architecture of the victim model as the surrogate model. The results show that our strategies are effective in boosting clean-label attacks on face recognition tasks, posing a serious security threat. Selecting samples with self-supervised models increases the success rate of clean-label attacks significantly, showing the effectiveness of self-supervised features. On the other hand, since there is a high distribution shift from ImageNet and TinyImageNet to PubFig, the improvements of Supervised \Pretrained~and \OOD~are not as high as Self-supervised \Pretrained.
\begin{table}[ht!]
    \centering
    \footnotesize
    \caption{The attack success rate of our strategies with SIG on ResNet18/PubFig with $20\%$ and $50\%$ poisoning rate.}
    \begin{tabular}{lcc}
\toprule
Method                 & 20\%  & 50\%  \\ \midrule
Random                 & 13.20 & 29.31 \\
Self-supervised models & \bf 24.92 & \bf 62.09 \\
Supervised models      & 19.26 & 40.37 \\
Single-class OOD       & 16.27 & 48.27 \\ \bottomrule
\end{tabular}
    
    \label{tab:pubfig}
\end{table}

\subsection{The Number of Classes in the OOD Dataset}~\label{sec:ablation}
\begin{table}[ht!]
\begin{minipage}{0.45\textwidth}
\scriptsize
\setlength{\tabcolsep}{2pt}
\renewcommand{\arraystretch}{1.0}
\centering
\caption{ASR of Multiple-class \OOD~ when varying the number of classes in the OOD dataset.}
\begin{tabular}{llcccccc}
\toprule
\multirow{2}{*}{Dataset} & \multirow{2}{*}{\begin{tabular}{@{}l}Number of \\labels\end{tabular}} & \multicolumn{3}{c}{ResNet18} & \multicolumn{3}{c}{VGG19} \\ \cmidrule(lr){3-5}\cmidrule(lr){6-8} 
 &  & 5\% & 10\% & 20\% & 5\% & 10\% & 20\% \\ \midrule
\multirow{2}{*}{CIFAR10} & 10 & 65.11 & 80.76 & 88.79 & 50.81 & 65.80 & 78.28 \\
 & 200 & 65.89 & 71.26 & 77.18 & 39.13 & 54.62 & 66.20 \\ \midrule
 \multirow{2}{*}{GTSRB} & 10 & 43.56 & 47.59 & 52.50 & 27.93 & 28.98 & 31.94 \\
 & 200 & 46.39 & 48.36 & 54.00 & 23.26 & 27.09 & 24.32 \\
 \bottomrule
\end{tabular}
\label{tab:abla_mutliple}
\end{minipage}\hfill
\begin{minipage}{0.45\textwidth}
\scriptsize
\setlength{\tabcolsep}{2pt}
\renewcommand{\arraystretch}{1.0}
\centering
\caption{ASR of Single-class \OOD~ when varying the number of classes in the OOD dataset.}
\begin{tabular}{llcccccc}
\toprule
\multirow{2}{*}{Dataset} & \multirow{2}{*}{\begin{tabular}{@{}l@{}}Number of \\labels\end{tabular}} & \multicolumn{3}{c}{ResNet18} &\multicolumn{3}{c}{VGG19} \\ \cmidrule(lr){3-5}\cmidrule(lr){6-8} 
 &  & \multicolumn{1}{c}{5\%} & \multicolumn{1}{c}{10\%} & \multicolumn{1}{c}{20\%} & \multicolumn{1}{c}{5\%} & \multicolumn{1}{c}{10\%} & \multicolumn{1}{c}{20\%} \\ 
 \midrule
 \multirow{2}{*}{CIFAR10} & 10 & 72.93 & 79.07 & 87.18 & 57.24 & 72.35 & 79.04 \\
 & 200 & 69.74 & 81.12 & 86.40 & 50.33 & 68.46 & 80.28 \\
  \midrule
\multirow{2}{*}{GTSRB} & 10 & 49.07 & 51.71 & 55.15 & 35.91 & 38.54 & 38.28 \\
 & 200 & 47.34 & 50.04 & 55.52 & 35.85 & 42.33 & 39.03 \\ \bottomrule
\end{tabular}
\label{tab:abla_single}
\end{minipage}
\end{table}

We study the performance for the \OOD~with the different number of classes in the OOD dataset. The results of Multiple-class \OOD~using SIG triggers are illustrated in Table~\ref{tab:abla_mutliple}. 
In general, these observations suggest that increasing the number of data labels in the OOD dataset does not improve the attack effectiveness.

\subsection{Effectiveness of Our Method on Narcissus} 
Similar to our method, the Narcissus attack \cite{zeng2023narcissus} can also operate under the threat model scenario where the attacker only has access to training data from the target class. 
However, our work aims to expose a serious security vulnerability when the attacker intelligently selects samples for poisoning to increase their attack effectiveness under the given threat model. This sample selection strategy allows the attacker to use any existing triggers in the clean-label setting. 
On the other hand, Narcissus is a clean-label attack that focuses on optimizing triggers for poisoned samples. 
In other words, our work is orthogonal to Narcissus since the proposed sample selection method can be used in Narcissus to intelligently (instead of randomly) select samples for poisoning to achieve better attack effectiveness, similar to what we demonstrate for BadNet, SIG, or Blended. To verify this, we perform experiments where the base attack is Narcissus and assess its performance with different sample selection approaches, including random and easy or hard samples found by a self-supervised model, to poison 25 samples on CIFAR10. Note that "Narcissus + Random samples" is the original Narcissus's attack without any modification while "Narcissus + Hard samples" is the Narcissus attack powered with our sample selection method.

In Table \ref{tab:appendix-combine-with-narcissus}, we report the mean success rate of three times attacking a ResNet18 model using the Narcissus approach with different sample selection strategies. As we can observe, the unmodified Narcissus's attack can only achieve 56.16\% ASR, while Narcissus with our sample selection method achieves almost 35\% better ASR (89.65\% ASR). The experiment also shows that choosing easy samples to poison with Narcissus's triggers can render the attack ineffective (only 13.06\% ASR). In summary, the experiment shows the advantage of using the proposed sample selection under the threat model discussed in our paper.

\begin{table}[ht!]
\centering
\footnotesize
\caption{Narcissus's performance with different sample selection approaches}
\label{tab:appendix-combine-with-narcissus}
\begin{tabular}{lc}
\toprule
                             & ASR   \\
\midrule
Narcissus + Easy samples     & 13.06          \\
Narcissus + Random selection & 56.16          \\ 
Narcissus + Hard samples     & \textbf{89.65} \\ 
\bottomrule
\end{tabular}
\end{table}




\subsection{The Transferability of Our Method to Different Architecture} We provide additional results for other architecture in Table~\ref{tab:additional_architectures}. We perform clean-label backdoor attacks on ViT~\cite{dosovitskiy2021an} or DeiT~\cite{touvron2021training} with 500 samples selected by a self-supervised pre-trained ResNet50 model~\cite{resnet}. As can be observed, our strategy improves the attack success rate of the random selection strategy by a large margin (more than $30\%$ for BadNet and $10\%$ for Blended and SIG), demonstrating that the selected samples still help to achieve effective attacks on various architectures.



\begin{table}[ht!]
\centering
\footnotesize
\caption{The performance of \Pretrained~on ViT/CIFAR10 and DeiT/GTSRB models}
\label{tab:additional_architectures}
\begin{tabular}{llcccccc}
\toprule
& \multirow{2}{*}{Method} & \multicolumn{2}{c}{BadNet} & \multicolumn{2}{c}{Blended} & \multicolumn{2}{c}{SIG} \\
\cmidrule(lr){3-4} \cmidrule(lr){5-6} \cmidrule(lr){7-8}
& & ACC & ASR & ACC & ASR & ACC & ASR \\
\midrule
\multirow{2}{*}{ViT/CIFAR10}
& Random & 98.78 & 11.26 & 98.80 & 23.10 & 98.72 & 35.55 \\
& Pretrained strategy & 98.84 & 47.20 & 98.79 & 38.63 & 98.73 & 45.00 \\
\midrule
\multirow{2}{*}{DeiT/GTSRB} & 
Random & 97.85 & 6.66 & 98.42 & 26.90 & 98.04 & 31.86 \\
& Pretrained strategy & 97.66 & 7.49 & 97.91 & 54.66 & 97.91 & 38.31 \\
\bottomrule
\end{tabular}
\end{table}

\subsection{The Effect of Class Imbalance} To study the effect of our method and class imbalance, we sort the classes by the number of samples descendingly and launch the backdoor attacks (with SIG trigger and 10\% poisoning rate) on the classes at 1st, 14th, 28th, and 43rd sorted positions (as target classes), whose original class indices are 1, 11, 16, and 37, respectively. The results in Table \ref{tab:class_imbalance} show that our method can consistently boost the success rates of the original attacks (random selection) on target classes from a broad spectrum of sample sizes.
\begin{table}[ht!]
\centering
\footnotesize
\caption{The performance of \Pretrained~on classes with different sizes.}
\label{tab:class_imbalance}
\begin{tabular}{lcccc}
\toprule
 & 1st & 14th & 28th & 43rd \\
\midrule
Number of samples & 1500 & 900 & 300 & 150 \\
ASR with random selection & 48.07 & 43.87 & 1.66 & 16.30 \\
ASR with Pretrained strategy & 57.12 & 44.70 & 24.36 & 27.75 \\
\bottomrule
\end{tabular}
\end{table}

\subsection{The Effect of Different Values of $k$} One of our approaches employs $k$-NN to select samples that have the highest distances to their neighbors in the feature space. We vary the number $k$ of neighbors and perform SIG attack with $10\%$ poisoning rate on CIFAR10 and report the success rate in Table \ref{tab:appendix-vary-k}. The results imply that $k$-NN is more effective when the value of $k$ is small; we conjecture that $k$NN with smaller $k$ takes into account the local property of the dataset, increasing the discrepancy between the score of hard samples (which are outliers) and easy samples. In the extreme case where $k=10000$ (meaning we select samples that are far from the mean), while the success rate is still higher than that of random selection, it is lower than that of a smaller k. Consequently, this suggests the use of a small k value when performing the attacks.
\begin{table}[ht!]
\centering
\footnotesize
\caption{The performance of \Pretrained~with different values of $k$.}
\label{tab:appendix-vary-k}
\begin{tabular}{lccccccc}
\toprule
& k=5 & k=50 & k=500 & k=1000 & k=5000 & k=10000 & Random \\
\midrule
ASR & 82.35 & 80.59 & 78.92 & 79.41 & 77.34 & 74.76 & 60.54 \\
\bottomrule
\end{tabular}
\end{table}

\subsection{Access only a portion of the data from the target class.} We study the effectiveness of our method under the setting where the attacker partially accesses the target class's data. We conducted experiments using different selection strategies on CIFAR-10 with accessible data comprising $20\%$ and $50\%$ subsets of the target class's data (Table \ref{tab:appendix-accessible-data-sizes}). The attack used was SIG, and the poisoning rate was $5\%$. We can observe that as the size of the accessible data decreases, the corresponding ASR also decreases. However, the method's effectiveness is still significantly better than that of random selection. The attackers, using our method, can still launch a very harmful attack ($68.92\%$) even with only $20\%$ of the target class's data.

\begin{table}[ht!]
\centering
\footnotesize
\caption{The performance of \Pretrained~with partial access to the target class.}
\label{tab:appendix-accessible-data-sizes}
\begin{tabular}{lcccccc}
\cmidrule(l){2-7}
 & \multicolumn{2}{c}{20\%} & \multicolumn{2}{c}{50\%} & \multicolumn{2}{c}{100\%} \\ \cmidrule(lr){2-3} \cmidrule(lr){4-5}  \cmidrule(lr){6-7} 
 & Acc & ASR & Acc & ASR & Acc & ASR \\
Random & 94.63 & 45.27 & 94.80 & 43.37 & 94.69 & 50.28 \\
Pretrained strategy & 94.44 & 68.92 & 94.73 & 70.65 & 94.71 & 76.35 \\ \bottomrule
\end{tabular}
\end{table}

\section{Limitations}~\label{sec:limitation}
As discussed, we study the threat model in which the attacker acts as a data provider that is responsible for a single data class and we propose strategies to select samples for more effective clean-label attacks. We do not include dirty-label attacks that change the semantic label of the input since it is easier to detect in this threat model. We also do not study the effect of our strategies on the defense that mitigates the backdoor in training data.

Our methods select hard samples in the target class to inject trigger, and it would be interesting to study the combination of our strategies and the attack that perturbs the input to make it difficult to learn~\cite{turner2019label}. We leave these to future works.

\section{Societal Impacts}\label{sec:impact}
Our work proposes a novel threat model, where the adversary only has access to the target class that they want to attack. 
In this constrained setting, we show that the attacker can perform selective poisoning to improve the attack success rate of existing clean-label attacks. 
We hereby raise awareness of a new potential risk in developing a machine learning system in practice.
\end{document}